\documentclass[sigconf]{acmart}
\AtBeginDocument{%
  \providecommand\BibTeX{{%
    \normalfont B\kern-0.5em{\scshape i\kern-0.25em b}\kern-0.8em\TeX}}}

\setcopyright{acmlicensed}
\copyrightyear{2023}
\acmYear{2023}
\acmDOI{10.1145/3581783.3611774}

\acmConference[MM '23] {Proceedings of the 31st ACM International Conference on Multimedia}{October 29--November 3, 2023}{Ottawa, ON, Canada.}
\acmBooktitle{Proceedings of the 31st ACM International Conference on Multimedia (MM '23), October 29--November 3, 2023, Ottawa, ON, Canada}
\acmPrice{15.00}
\acmISBN{979-8-4007-0108-5/23/10}

\usepackage{algorithm}
\usepackage{algorithmic}
\usepackage{amsmath}
\usepackage{makecell}
\usepackage{array}
\usepackage{multirow}
\usepackage{booktabs}
 \usepackage{enumitem}
\newcommand{\etal}{\textit{et al}.}
\newcommand{\ie}{\textit{i}.\textit{e}.}
\newcommand{\eg}{\textit{e}.\textit{g}.}

\newcommand{\whzjh}[1]{{\color{black}#1}}

\begin{document}

\title{Prompted Contrast with Masked Motion Modeling: Towards Versatile 3D Action Representation Learning}

\author{Jiahang Zhang}
\affiliation{%
  \institution{Wangxuan Institute of Computer Technology, Peking University}
  \city{Beijing}
  \country{China}
}
\email{zjh2020@pku.edu.cn}

\author{Lilang Lin}
\affiliation{%
  \institution{Wangxuan Institute of Computer Technology, Peking University}
  \city{Beijing}
  \country{China}
  }
\email{linlilang@pku.edu.cn}

\author{Jiaying Liu}
\authornote{Corresponding author. This work is supported by the National Natural Science Foundation of China under contract No.62172020.}
\affiliation{%
  \institution{Wangxuan Institute of Computer Technology, Peking University}
  \city{Beijing}
  \country{China}
  }
\email{liujiaying@pku.edu.cn}

\renewcommand{\shortauthors}{Jiahang Zhang, Lilang Lin, \& Jiaying Liu.}

\begin{abstract}
Self-supervised learning has proved effective for skeleton-based human action understanding, which is an important yet challenging topic. 
Previous works mainly rely on contrastive learning or masked motion modeling paradigm to model the skeleton relations. However, the sequence-level and joint-level representation learning cannot be effectively and simultaneously handled by these methods. As a result, the learned representations fail to generalize to different downstream tasks. Moreover, combining these two paradigms in a naive manner leaves the synergy between them untapped and can lead to interference in training. To address these problems, we propose \textbf{P}rompted \textbf{C}ontrast with \textbf{M}asked \textbf{M}otion \textbf{M}odeling, PCM$^{\rm 3}$, for \textit{versatile} 3D action representation learning. Our method integrates the contrastive learning and masked prediction tasks in a mutually beneficial manner, which substantially boosts the generalization capacity for various downstream tasks. Specifically, masked prediction provides novel training views for contrastive learning, which in turn guides the masked prediction training with high-level semantic information. Moreover, we propose a dual-prompted multi-task pretraining strategy, which further improves model representations by reducing the interference caused by learning the two different pretext tasks.
Extensive experiments on \textit{five} downstream tasks under three large-scale datasets are conducted, demonstrating the superior generalization capacity of PCM$^{\rm 3}$ compared to the state-of-the-art works. Our project is publicly available at: {https://jhang2020.github.io/Projects/PCM3/PCM3.html}.
\end{abstract}



\maketitle

\begin{figure}[t]
    \centering
    \includegraphics[width=0.49\textwidth]{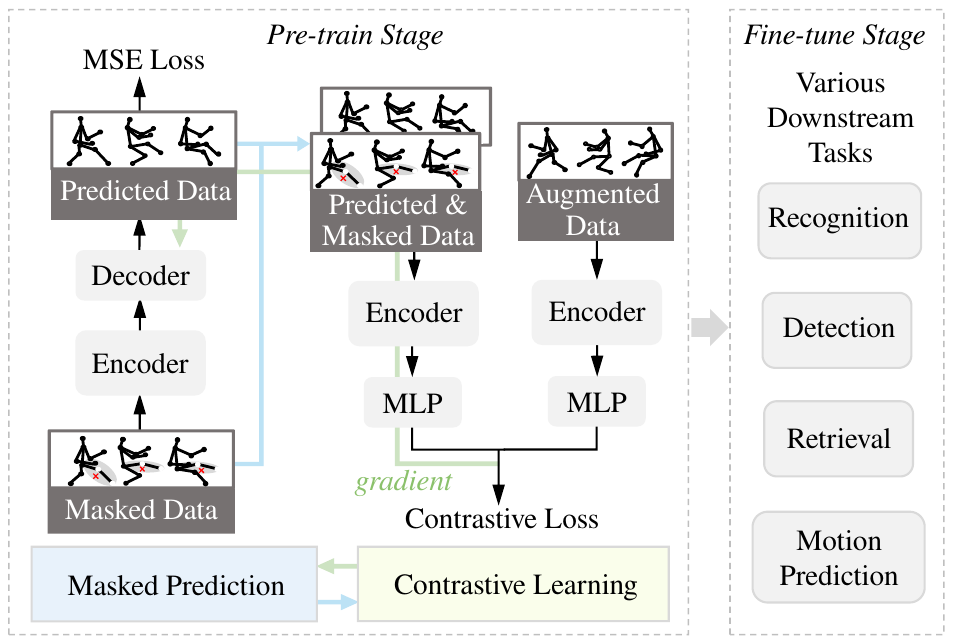}
   \caption{
   Illustration of the proposed method for versatile action representation learning. We integrate contrastive learning and masked skeleton modeling paradigms in a mutually beneficial manner. The masked prediction provides novel views for contrastive learning (blue arrow), and the generated gradients of contrast (green arrow) serve as high-level semantic guidance for masked prediction in turn, modeling both joint-level and sequence-level features.
   }
   \vspace{-5pt}
  \label{fig:teaser}
\end{figure}

\section{Introduction}
Human activity understanding is a crucial problem in multi-media processing on account of its significant role in real-life applications, such as human-robotics interaction~\cite{lee2020real}, healthcare~\cite{lopez2019human} and autonomous driving~\cite{camara2020pedestrian}.
As a highly efficient representation for human activity understanding, 3D skeletons represent the human form by 3D coordinates of key body joints. In comparison to other representations such as RGB videos and depth data, skeletons are lightweight, compact, and privacy-preserving. Owing to these competitive advantages, skeletons have become widely used in human action analysis.

Many efforts have been made on the supervised skeleton-based action learning~\cite{du2015hierarchical,yan2018spatial,shi2019two,cheng2020skeleton}. However, the performance of these methods heavily relies on huge amount of labeled data, which requires time-consuming and expensive data annotation work. This inherent shortcoming of full supervision limit their applications in the real world. Therefore, more and more attention has been paid to self-supervised 3D action representation learning recently to learn meaningful features from unlabeled data.

Self-supervised 3D action representation learning research has mainly focused on two paradigms: reconstruction-based and contrastive learning-based methods. Reconstruction-based methods leverage an encoder-decoder architecture to learn representations by predicting masked skeletons (\ie,~masked modeling) or reconstructing original data. These methods focus on joint-level feature modeling and capture spatial-temporal relationships. In contrast, contrastive learning-based methods use data augmentations to construct positive/negative pairs, and apply an instance discrimination task to learn sequence-level semantic features.

However, it is noticed that most recent representation learning methods focus on the single paradigm to model the joint-level (by masked skeleton modeling)~\cite{zheng2018unsupervised,wu2022skeletonmae} or sequence-level (by contrastive learning)~\cite{li20213d,mao2022cmd,hico2023,zhang2022hiclr} features solely. As a result, it is difficult for these methods to generalize well to different downstream tasks, \eg,~recognition task and motion prediction, because they cannot learn representations of different granularity simultaneously and effectively.
Although some works~\cite{lin2020ms2l,su2020predict,wang2022contrast} make valuable efforts to combine the above two approaches to learn richer representations, only mediocre improvement is observed. It is because simply combining them ignores the interference due to the gaps between feature modeling mechanisms of masked prediction and contrastive learning~\cite{qi2023contrast}, and fails to utilize the potential synergy.
These problems limit the generalization power of model, and versatile 3D action representation learning remains a challenging and under-explored area.

To this end, we propose the \textit{prompted contrast with masked motion modeling}, PCM$^{\rm 3}$, which explores the mutual collaboration between the above two paradigms for versatile 3D action representation learning as shown in Figure~\ref{fig:teaser}. Specifically, the well-designed inter- intra- contrastive learning and topology-based masked skeleton prediction are first proposed as the basic pipelines. Furthermore, we connect the two tasks and explore the synergy between them. The views in masked prediction training are utilized as novel positive samples for contrastive learning. 
In turn, the masked prediction branch is also updated via the gradients from the contrastive learning branch for higher-level semantic guidance. Meanwhile, to reduce the distraction of learning between different pretext tasks and data views, we propose the dual-prompted multi-task pretraining strategy. Two types of prompts, namely, domain-specific prompts and task-specific prompts are applied to explicitly instruct the model to learn from different data views/tasks. 
Extensive experiments under \textit{five} downstream tasks are conducted to provide a comprehensive evaluation. The proposed method demonstrates promising generalization capacity compared to state-of-the-art methods.
Our contributions can be summarized as follows:
\begin{itemize}[leftmargin=2em]

\item We propose PCM$^{\rm 3}$ for multi-granularity representations, which integrates masked skeleton prediction and contrastive learning paradigms in a mutually beneficial manner. We employ the masked prediction network to generate more diverse positive motion views for contrastive learning. Meanwhile, the generated gradients are propagated and guide masked prediction learning in turn with high-level semantic information.

\item Considering that different data views and pretraining tasks can cause mutual interference, we introduce domain-specific prompts and task-specific prompts for the multi-task pretraining. These trainable prompts enable the model to achieve more discriminative representations for different skeletons.

\item We perform rigorous quantitative experiments to assess the generalization efficacy of state-of-the-art self-supervised 3D action representation learning techniques across five downstream tasks, including recognition, retrieval, detection, and motion prediction, on both uncorrupted and corrupted skeletons. Our study serves as a comprehensive benchmark for the research community, and we believe it can provide valuable insights and aid future investigation in this field.

\end{itemize}

\section{Related Works}
\subsection{Skeleton-based Action Recognition}
With the huge advances of deep learning, \whzjh{recurrent neural network (RNN)}-based, \whzjh{convolutional neural network (CNN)}-based, graph convolutional network (GCN)-based\whzjh{,} and \whzjh{transformer}-based methods are studied for skeleton-based action recognition.
RNNs have been widely used to model temporal dependencies and capture the motion features for skeleton-based action recognition. 
\whzjh{The work in}~\cite{du2015hierarchical} uses RNN to tackle the skeleton as \whzjh{sequence} data. 
Subsequently, Song \etal~\cite{song2017end,song2018skeleton} proposed to utilize the attention mechanism and multi-modal information to enhance the feature representations.
Some other works~\cite{ke2017new,liu2017enhanced} transform each skeleton sequence into \whzjh{image-like} representations and apply the CNN model to extract spatial-temporal information. 
Recently, GCN-based methods have attracted more attention due to the natural topology structure of the human body.
Many works~\cite{yan2018spatial,shi2019two,cheng2020skeleton} apply GCN to the spatial and temporal dimension ~\cite{yan2018spatial} and achieves remarkable results in the supervised skeleton-based action recognition.
Meanwhile, \whzjh{transformer} models~\cite{shi2020decoupled,plizzari2021skeleton} also show promising results, owing to long-range temporal dependency learning by attention. 

However, these supervised works rely on the huge labeled data to train the model. In this paper, we explore the self-supervised 3D action representation learning instead.
\vspace{-2pt}
\subsection{Contrastive Learning for Skeleton}
Contrastive learning~\cite{he2020momentum,chen2020simple,chen2020improved} has proven effective for skeleton representation learning.
One popular research view is the skeleton augmentations, which are crucial for the learned representation quality.
Guo~\etal~\cite{guo2022aimclr} explored the use of extreme augmentations in the current contrastive learning pipeline. Zhang~\etal~\cite{zhang2022hiclr} proposed hierarchical consistent contrastive learning to utilize more strong augmentations. 
Another perspective is to explore the knowledge of different views in the skeleton. ISC~\cite{thoker2021skeleton} performs a cross-contrastive learning manner using image, graph and sequence representations. Li {\etal}~\cite{li20213d} mined the potential positives resort to the different skeleton modalities, \ie,~joint, bone, motion, and re-weights training samples according to the similarity. Mao~\etal~\cite{mao2022cmd} performed the mutual-distillation across different views. 
Different from the above works, we propose to integrate the masked modeling pretext task with contrastive learning, modeling both joint-level and sequence-level features for more general representation learning. 

\subsection{Masked Image/Skeleton Modeling}
Masked modeling has been explored in stacked denoising autoencoders~\cite{vincent2010stacked}, where the mask operation is regarded as adding noise to the original data. 
Recently, the masked modeling has achieved remarkable success in self-supervised learning~\cite{he2022masked,xie2022simmim} for image representation learning. 
For skeleton data, LongT GAN~\cite{zheng2018unsupervised} directly utilizes an autoencoder-based model optimized by an additional adversarial training strategy. 
Some works~\cite{su2020predict,lin2020ms2l} apply the motion prediction pretext task to learn the temporal dependencies in skeleton sequences. Inspired by the masked autoencoder~\cite{he2022masked}, Wu~\etal~\cite{wu2022skeletonmae} proposed a masked skeleton autoencoder to learn the spatial-temporal relationships. 
In this paper, we explore the synergy between masked modeling and contrastive learning and propose a topology-based masking strategy to further boost the representation learning.

\section{The Proposed Method: PCM$^{\rm 3}$}
In this part, we firstly describe our designed pipelines for contrastive learning (Section~\ref{sec:cl}) as well as our proposed topology-based method for masked modeling (Section~\ref{sec:mp}). Then, we further present the synergy exploration between the two pretext tasks in Section~\ref{sec:syn}. The prompt-based pretraining strategy and the whole model are given in Section~\ref{sec:prompt}.

\begin{figure*}[tb]
 \centering
  \includegraphics[width=0.99\textwidth ]{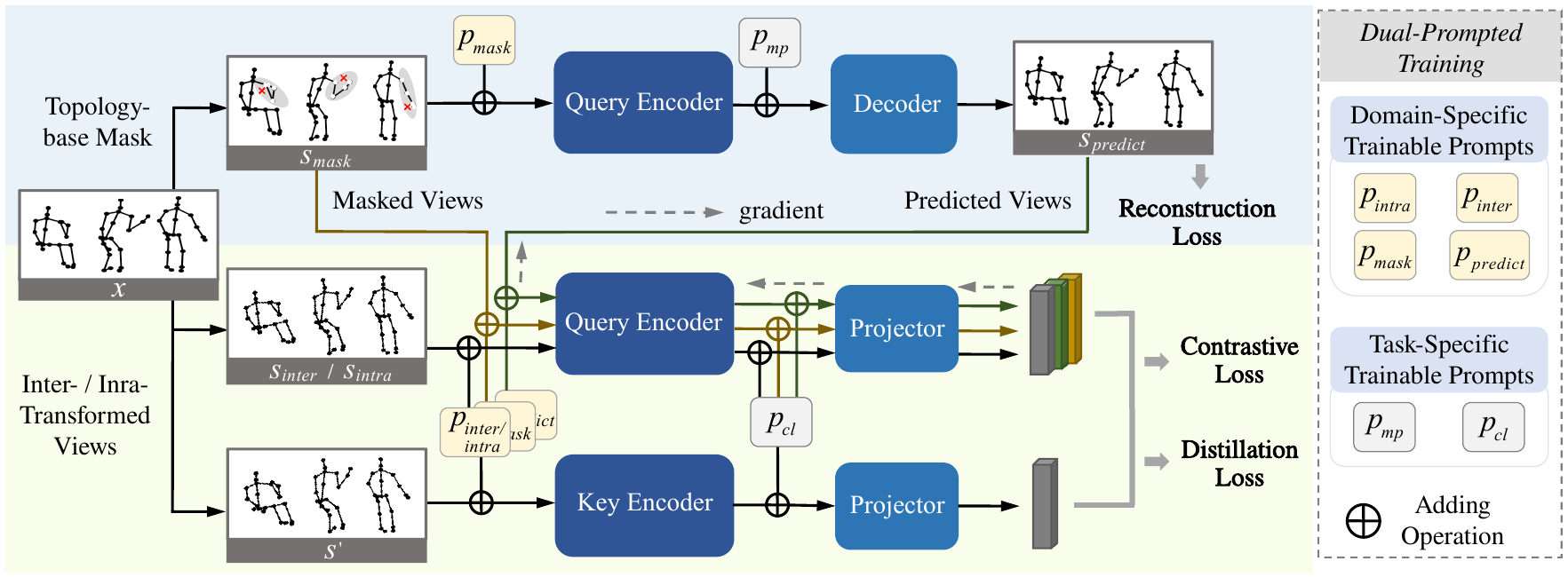}
  \caption{The overview of the proposed method. We integrate the masked skeleton prediction (\textit{blue} part) and the contrastive learning (\textit{yellow-green} part) paradigms in a mutually beneficial manner. For brevity, we represent intra- and inter- transformed views in a single branch in the diagram.
 The masked and the predicted views are utilized to expose more novel motion patterns for contrastive learning. Meanwhile, the gradients from contrastive learning (\textit{dotted arrows} in figure) are propagated to the masked prediction branch to update the decoder. To further boost the representation learning from different views/tasks, we propose the dual-prompted multi-task pretraining strategy, where domain-specific and task-specific prompts are added in input-wise and feature-wise form, serving as training guidance.}
 \label{fig:pip_inv}
\end{figure*}

\subsection{Skeleton Contrastive Learning}\label{sec:cl}
For clarity, the canonical design of contrastive learning for image/skeleton is given following previous works~\cite{he2020momentum,chen2020simple,bai2022directional}, which usually comprises the following components:
\begin{itemize}[leftmargin=2em]
\item \textbf{Data augmentation module} contains a series of manual data transformations to construct different views of original data, which are regarded as positive samples sharing the same semantic. 
\item \textbf{Encoder $f(\cdot)$} serves as the mapping function from the input space to the latent feature space.
\item \textbf{Embedding projector $h(\cdot)$} is successively applied after encoder $f(\cdot)$, mapping the encoded feature into an embedding space where the self-supervised loss is applied.
\item \textbf{Self-supervised loss} aims to maximize the similarity between positive samples, performing the feature clustering operation to obtain a distinguishable representation space.
\end{itemize}

Our contrastive learning design is based on MoCo v2~\cite{chen2020improved}. Specifically, we introduce intra- inter- skeleton transformations and relational knowledge distillation, to assist model to capture diverse motion patterns and boost the representation learning.

\vspace{0.5em}

\noindent\textbf{1) Intra-skeleton transformation learning.} We utilize the following transformations: \textit{Temporal crop-resize, Shear,} and \textit{Joint Jittering} following the previous works~\cite{thoker2021skeleton,mao2022cmd}. Specifically
given a skeleton sequence $x$, the positive pair $(s_{intra}, s^{\prime})$ is constructed via the above transformations. Then, we obtain the corresponding feature representations $(z_{intra}, z^{\prime})$ via the query/key encoder \whzjh{$f_q(\cdot)/f_k(\cdot)$} and embedding projector $h_q(\cdot)/h_k(\cdot)$, respectively. 
Meanwhile, a memory queue \textbf{M} is maintained storing numerous negative samples for contrastive learning. We optimize the whole network by InfoNCE objective~\cite{oord2018representation}:
\begin{equation}\label{eq:infonce}
	\mathcal{L}_{Info}^{Intra} = -\log\frac{\exp(z_{intra} \cdot z^{\prime} / \tau)}{\exp(z_{intra} \cdot z^{\prime} / \tau) + \sum_{i=1}{\exp(z_{intra} \cdot m_{i} / \tau)}},
\end{equation}
\whzjh{where $m_{i}$ is the feature in \textbf{M} corresponding to the $i$-th negative sample and $\tau$ is the temperature hyper-parameter.}
After a training step, the samples in a batch will be updated to \textbf{M} as negative samples according to a first-in, first-out policy.
The key encoder is a momentum-updated version of the query encoder, \ie,~$\theta_{k} \leftarrow \alpha\theta_{k} + (1-\alpha)\theta_{q},$
where $\theta_{q}$ and $\theta_{k}$ are the parameters of query encoder and key encoder, and $\alpha \in [0,1)$ is a momentum coefficient.

\vspace{0.5em}

\noindent\textbf{2) Inter-skeleton transformation learning.} Inspired by the successful application of \textit{Mix} augmentation in self-supervised learning~\cite{kim2020mixco,lee2020mix,shen2022mix,chen2022skelemixclr}, we introduce the \textit{CutMix}~\cite{yun2019cutmix}, \textit{ResizeMix}~\cite{ren2022simple}, and \textit{Mixup}~\cite{zhang2017mixup} to our skeleton contrastive learning. These inter-skeleton transformations utilize two different samples to generate mixed augmented views. Specifically, given two skeleton sequences $s_1, s_2$, we randomly select a mixing method from the above and obtain the mixed skeleton data $s_{inter}$ as follows:

\noindent $\bullet$ \textit{Mixup}~\cite{zhang2017mixup}: We interpolate the two skeleton sequences according to a sampled mixing ratio $\lambda$, \ie,~$s_{inter} = (1-\lambda) s_1 + \lambda s_2$.

\noindent $\bullet$ \textit{CutMix}~\cite{yun2019cutmix}: The randomly selected regions of two skeleton sequences are cut and pasted across the spatial-temporal dimension. And $\lambda$ is defined as the ratio of replaced joint number to the total joint number.

\noindent $\bullet$ \textit{ResizeMix}~\cite{ren2022simple}: This is similar to \textit{CutMix}, but downsamples $s_{2}$ first in the temporal dimension before mixing. 

Subsequently, we can obtain the embeddings corresponding to the mixed data by $z_{inter} = h_q\circ f_q(s_{inter})$, and the following loss is optimized:
\begin{equation}
	\mathcal{L}_{Info}^{Inter} = -\log\frac{\exp(z_{inter} \cdot z_{inter}^{\prime} / \tau)}{\exp(z_{inter} \cdot z_{inter}^{\prime} / \tau) + \sum_{i=1}{\exp(z_{inter} \cdot m_{i} / \tau)}},
\end{equation}
where $z_{inter}^{\prime} = (1-\lambda) (h_{k}\circ f_k(s_1)) + \lambda (h_{k}\circ f_k(s_2))$.

\vspace{0.5em}
\noindent\textbf{3) Relational Knowledge Distillation.} To further provide fine-grained semantic consistency supervision for contrastive learning,  we introduce a relational knowledge self-distillation loss to positive pairs. Inspired by the works~\cite{wei2020co2,zhang2022hiclr,mao2022cmd}, the relational knowledge is modeled as the cosine similarity between $z^{\prime}$/$z^{\prime}_{inter}$ and feature anchors in memory queue $\textbf{M}$. The relational distribution, \ie, the similarity with respect to negative anchor samples, is enforced to be consistent between each positive pair. Taking the embedding pair $(z_{intra},z^{\prime})$ corresponding to the aforementioned intra-transformation as an example, the loss can be expressed as:
\begin{equation}\label{eq:kd}
\begin{aligned}
    \mathcal{L}^{Intra}_{KL} &= -{p}\left({z}^{\prime},\tau_{k}\right)\log{{p}\left({z_{intra}},\tau_{q}\right)}, \\{p_{j}}\left({z},\tau\right) &= \frac{\exp({z} \cdot {m}_{j} / \tau)}{\sum_{i=1}{\exp({z} \cdot {m}_{i} / \tau)}},
\end{aligned}
\end{equation}
where $m_{i}$ is the stored $i_{th}$ feature anchors in \textbf{M}. $\tau_{k}$ and $\tau_{q}$ are temperature hyper-parameters, set as 0.05 and 0.1, respectively. 
This distillation term introduces more anchors to mine the fine-grained and semantics-aware similarity relations~\cite{wei2020co2}, boosting the representation quality. 

\begin{figure}[t]
   \centering
   \includegraphics[width=0.47\textwidth]{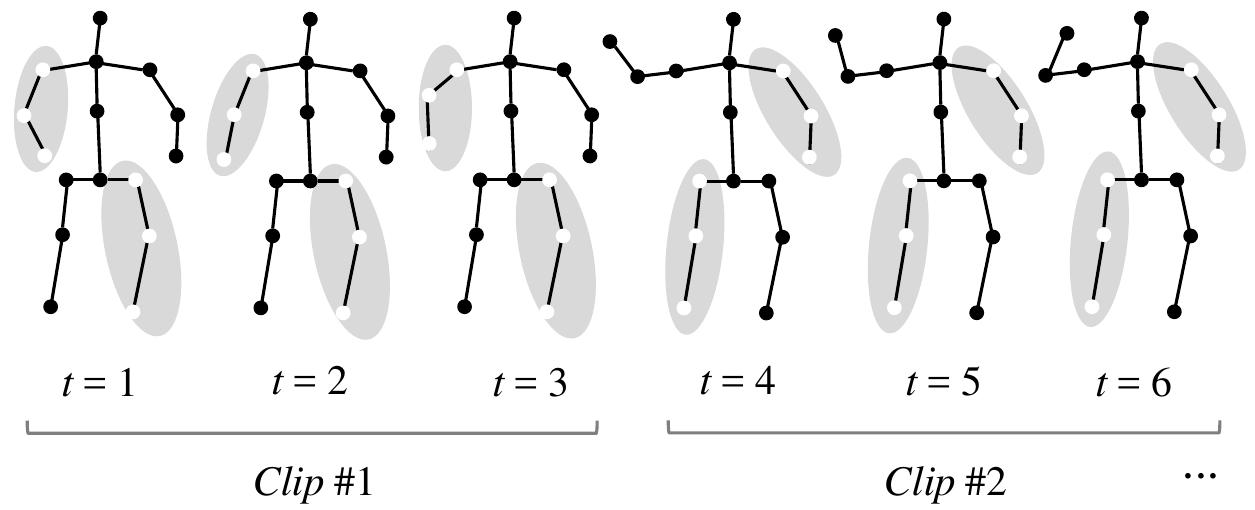}
   \caption{Illustration of the topology-based masking strategy. The gray region is the body parts to be masked.}
  \label{fig:topo_mask}
\end{figure}

\subsection{Masked Skeleton Prediction}\label{sec:mp}
To further enrich the learned representations by model, we integrate masked skeleton modeling and the joint-level feature learning is baked into training process. This further improves generalization ability especially for dense prediction downstream tasks, as compared with using only instance-wise discrimination task, \ie, contrastive learning.

First, in terms of the masking strategy, previous works~\cite{zheng2018unsupervised,wu2022skeletonmae} utilize \textit{Random Mask} to randomly select the masked joints in spatial-temporal dimension. However, given the redundancy of skeleton sequences, the masked joints can be easily inferred by copying adjacent joints in spatial or temporal dimension, which is not conducive to modeling meaningful relationships in skeletons. To this end, we propose \textit{Topology-based} masking strategy, which masks the skeleton in the body-part level instead of the joint level, \ie~\textit{trunk, right-hand, left-hand, right-leg} and \textit{left-leg}. Meanwhile, we divide the sequences into different clips in temporal dimension, and the same parts are all masked in a clip, as shown in Figure~\ref{fig:topo_mask}.

Based on the above masking strategy, we mask the original skeleton $x$ and then feed the masked skeleton $s_{mask}$ into the encoder $f_q(\cdot)$ to obtain the corresponding features. To predict the masked skeleton region, we employ a decoder $dec(\cdot)$, which takes the encoded features as input and outputs the reconstructed skeleton. The MSE loss between original data $x$ and predicted data $s_{predict}$ is optimized in the masked region:
\begin{equation}
\label{eq:mask}
\mathcal{L}_{Mask} = \frac{1}{N}\sum^{N}||(x - dec\circ f_q(s_{mask}))\odot (\mathit{1}-M) ||_{2},
\end{equation}
where $N$ is the number of all masked joints. 
$M$ is the binary mask in which $1$ and $0$ correspond to the visible joints and the masked joints, respectively. And $\mathit{1}$ is an all-one matrix with the same shape as $M$.

\subsection{Collaboration between Contrastive Learning and Masked Modeling}\label{sec:syn}
Despite our proposed novel pipelines for contrastive learning and masked skeleton prediction, we find that simply integrating the two paradigms only yield mediocre performance gains as shown in Table~\ref{tab:single_multi}. This is due to the inherent gap between feature modeling paradigm of the two tasks~\cite{qi2023contrast}, and the model cannot directly take advantage of the potential synergy between them.
Therefore, in this part, we explore the collaboration and connect the two tasks in a mutually beneficial manner.

\vspace{0.5em}

\noindent\textbf{1) Novel Positive Pairs as Connection.} First, we utilize special data views in masked prediction training to provide more diverse positive samples for contrastive learning. Considering that the masked skeleton naturally simulates occlusion for skeletons, we take masked skeleton $s_{mask}$ views as challenging positives, to learn the underlying semantic consistency and enhance robustness to occlusion. Meanwhile, 
we also boost contrastive learning by taking the predicted skeletons output by decoder $dec(\cdot)$ as positive samples. Compared with the masked views, the predicted views contain the inherent noise, uncertainty, and diversity brought by continuous training of the model, which contributes to encoding more diverse movement patterns and thus improves generalization capacity.

In a nutshell, we utilize the masked view $s_{mask}$ and predicted view $s_{predict}$ as positive samples to connect masked modeling with contrastive learning. We present all positive (embedding) pairs as follows:
\begin{equation}
\label{eq:pos}
\{(z_{intra}, z^{\prime}), (z_{inter}, z^{\prime}_{inter}), (z_{mask}, z^{\prime}), (z_{predict}, z^{\prime}) \}.
\end{equation}
They are obtained by the query/key encoder and projector, respectively. Each positive pair is applied to calculate Eq.~\ref{eq:infonce} for contrastive loss and Eq.~\ref{eq:kd} for distillation loss. Taking $(s_{mask},s^{\prime})$ as an example, it replaces the $z_{intra}$ with $z_{mask}=h_{q}\circ f_{q}(s_{mask})$ for optimization. \textbf{Note} that we use $\mathcal{L}_{Con}$ and $\mathcal{L}_{KL}$ to represent the total contrastive loss and distillation loss, respectively, which comprise component losses in the form of Eq.~\ref{eq:infonce} and Eq.~\ref{eq:kd} calculated for all four positive pairs defined in Eq.~\ref{eq:pos}.

\vspace{0.5em}

\noindent\textbf{2) High-Level Semantic Guidance.}
On the other hand, the gradients of $s_{predict}$ from the contrastive learning branch are propagated to update the reconstructed decoder $dec(\cdot)$ as shown in Figure~\ref{fig:pip_inv}. It provides the high-level semantic guidance for the skeleton prediction together with the MSE loss in Eq.~\ref{eq:mask} which serves as joint-level supervision, leading to better masked prediction learning and higher quality of $s_{predict}$ as positive samples.

With the above synergetic designs, the masked prediction task provides novel positive samples as meaningful supplements to the contrastive learning. Meanwhile, with the gradients of contrastive learning propagating to the masked modeling branch, the masked prediction task can be conversely assisted via the high-level semantic guidance provided by contrastive learning task. These designs connect the two tasks and yield better representation quality.

\subsection{Dual-Prompted Multi-Task Pretraining}\label{sec:prompt}
For self-supervised pretraining, the whole model is optimized for contrastive learning and masked prediction tasks in a multi-tasking manner. However, the input data are from different distributions (domains), \eg, augmented views and masked views, for different pretext tasks, \ie, contrastive learning and masked prediction. Previous works directly feed them into the encoder to learn respective representations. This can cause ambiguity and interfere with feature modeling in terms of learning from different data/tasks.

To this end, we propose a novel dual-prompted multi-task pretraining strategy to explicitly instruct the model to learn from different domains/tasks. Specifically, two types of prompts named \textit{domain-specific prompt} and \textit{task-specific prompt} are designed, which are implemented as trainable vectors to provide training guidance.

\vspace{0.5em}

\noindent\textbf{1) Domain-Specific Prompt}. To deal with different domains of input, we maintain domain-specific prompts for each input view, \ie, $p_{inter}, p_{intra}, p_{mask}$, and $p_{predict}$, of which the dimension equals to the skeleton spatial size. Then, these domain-specific prompts are added to the corresponding input data ($s_*$ means any view):
\begin{equation}
	s_{*} = s_{*} + p_{*}.
\end{equation}
These decorated skeletons are fed into the encoder for self-supervised pretraining.
The trainable prompts enable the model to learn domain-specific knowledge and achieve better representations. 

\vspace{0.5em}

\noindent\textbf{2) Task-Specific Prompt}. For task-specific prompts, we apply deep-feature prompt after encoder instead of input-wise prompt to encourage the encoder to extract more general features for various tasks. After obtaining the representations $feat_{*} = f_{q}/f_{k}(s_{*}) \in \mathbb{R}^{\rm d}$, we add the task-specific prompts $p_{cl}$ or $p_{mp}$ to the $feat_{*}$. Specifically, $p_{cl}, p_{mp} \in \mathbb{R}^{\rm r}$ where the dimension ${\rm r} < {\rm d}$ for efficiency~\cite{gan2022decorate}, are added to the randomly selected ${\rm r}$-dimensional channels from the original feature $feat_{*}$. If the feature is to feed into the projector $h_q/h_k(\cdot)$ for contrastive learning, the $p_{cl}$ is added, otherwise, the $p_{mp}$ is added for masked prediction.
These prompts can effectively learn task-specific knowledge and reduce interference between different pretext tasks.
 
Overall, the following objective is applied to the whole model as shown in Figure~\ref{fig:pip_inv}:
\begin{equation}
	\mathcal{L} = \mathcal{L}_{Con} + \lambda_{m}\mathcal{L}_{Mask} + \lambda_{kl}\mathcal{L}_{KL},
\end{equation}
where the loss weight $\lambda_{m}$ and $\lambda_{kl}$ are set to 40.0, 1.0 in implementation. Note that the prompts are tuned only in the pretraining stage since they are targeted for self-supervised pretext tasks rather than downstream tasks. Therefore, we simply drop all prompts after the pretraining stage.

\begin{table*}[t]
  \centering
  \caption{Comparison of unsupervised action recognition results.}
  \begin{tabular}{l|l|c|cc|cc|c}
    \toprule[0.8pt]
   \multirow{2}{*}{Method}&\multirow{2}{*}{Year}&\multirow{2}{*}{Backbone} &\multicolumn{2}{c|}{NTU 60}&\multicolumn{2}{c}{NTU 120}&PKUMMD\\
   & & &xsub (\%)& xview (\%)&xsub (\%)&xset (\%) &Part II (\%)\\
     \midrule
    {\textit{\color{black}{Single-stream:}}}\\
    Long TGAN~\cite{zheng2018unsupervised} &AAAI'2018 &\multirow{6}{*}{GRU} &39.1 &48.1  & - & - & 26.0\\
    MS$\rm^{2}$L~\cite{lin2020ms2l} &ACM MM'2020 & &52.6   &-    & - & - &27.6\\
    CRRL~\cite{wang2022contrast} &TIP'2022 & &67.6 & 73.8& 56.2 & 57.0 & 41.8\\
    ISC~\cite{thoker2021skeleton} &ACM MM'2021  & &76.3 &85.2 &67.1 &67.9 &36.0\\
    CMD~\cite{mao2022cmd} &ECCV'2022 & &79.8 &86.9 &70.3 &71.5 & 43.0\\
    HaLP~\cite{shah2023halp} &CVPR'2023 & &79.7 &86.8 &71.1 &72.2 &43.5\\
    \midrule
    H-Transformer~\cite{cheng2021hierarchical} &ICME'2021 &\multirow{2}{*}{Transformer} &69.3 &72.8 &- &- & -\\
    GL-Transformer~\cite{kim2022global} &ECCV'2022 & &76.3 &83.8 &66.0 &68.7 & -\\
    \midrule
    \textbf{PCM$^{\rm 3}$ (Ours)} &- &GRU &\textbf{83.9} & \textbf{90.4} &\textbf{76.5} &\textbf{77.5} & \textbf{51.5}\\
    \midrule[0.8pt]
    {\textit{\color{black}{Three-stream:}}}\\
    3s-HiCo~\cite{hico2023} &AAAI'2023 &\multirow{2}{*}{GRU}&82.6 &90.8 &75.9 &77.3 &-\\
    3s-CMD\cite{mao2022cmd} &ECCV'2022 &&84.1 &90.9 &74.7 &76.1 & 52.6\\
    \midrule
    3s-CrosSCLR~\cite{li20213d} &CVPR'2021&\multirow{6}{*}{GCN}&77.8 &83.4 & 67.9 & 66.7 &21.2\\
    3s-AimCLR~\cite{guo2022aimclr} &AAAI'2022& &78.9 &83.8 &68.2  & 68.8 &  38.5\\
    3s-HiCLR~\cite{zhang2022hiclr} &AAAI'2023& &80.4 &85.5 &70.0& 70.4 &53.8\\
    3s-HYSP~\cite{francohyperbolic} &ICLR'2023 & &79.1 &85.2 & 64.5 &67.3 &-\\
    3s-SkeleMixCLR~\cite{chen2022skelemixclr} &arXiv'2022& &82.7 & 87.1 & 70.5 & 70.7 & 57.1 \\ 
    3s-CPM~\cite{zhang2022cpm} &ECCV'2022& &83.2 &87.0 &73.0 &74.0 &51.5\\
    \midrule
    \textbf{3s-PCM$^{\rm 3}$ (Ours)} &- &GRU &\textbf{87.4} & \textbf{93.1} &\textbf{80.0} &\textbf{81.2} &\textbf{58.2}\\
     \bottomrule[0.8pt]
 \end{tabular}
 \label{tab:linear_ntu}
\end{table*}

\section{Experimental Results}

\subsection{Datasets}
\noindent\textbf{1) NTU RGB+D 60 Dataset (NTU 60)~\cite{shahroudy2016ntu}}. There are 56,578 videos with 25 joints in each frame. 60 action categories are defined in the dataset.
We adopt the following two evaluation protocols:  a) Cross-Subject (xsub): the data for training and testing are collected from 40 different subjects. b) Cross-View (xview): the data for training and testing are captured in 3 different views: front view, 45 degrees view for left side and right side.

\noindent\textbf{2) NTU RGB+D 120 Dataset (NTU 120)~\cite{liu2019ntu}}. NTU 120 is an extension to NTU 60 dataset. There are 114,480 videos collected with 120 action categories included. Two recommended protocols are also adopted: a) Cross-Subject (xsub): the data for training and testing are collected from 106 different subjects. b) Cross-Setup (xset): the data for training and testing are collected from 32 different setups with different camera locations.

\noindent\textbf{3) PKU Multi-Modality Dataset (PKUMMD)~\cite{liu2020pku}}. {PKUMMD is a large-scale dataset towards multi-modality 3D understanding of human actions. In particular, PKUMMD supports the evaluation of action detection.
The actions are organized into 51 action categories and almost 20,000 instances are included. The PKUMMD is divided into two subsets, Part I and Part II. We adopt cross-subject protocol following previous works.}

\subsection{Implementation Details}
We follow the experiment settings of the recent works~\cite{mao2022cmd,thoker2021skeleton}.
For data preprocessing, all sequences of skeletons are downsampled to 300 frames and then crop-resized to 64 frames to feed the model.
We adopt a 3-layer Bi-GRU as the encoder backbone, of which the hidden dimension is set as $d=1024$ following previous works. The task-prompt dimension $r$ is 128.
The MLPs are used as the projection heads, mapping features into embeddings with 128 dimensions. A 2-layer GRU with 512 dimensions is used as the decoder $dec(\cdot)$. 3s- denotes the fusion results of three streams, \ie~ joint, bone and motion modalities of skeleton.

During self-supervised pre-training, the model is trained for 450 epochs in total, with a batch size of 128. The initial learning rate is 0.02 and is reduced to 0.002 at 350$_{th}$ epoch. We employ the SGD optimizer with a momentum of 0.9 and the weight decay is 0.0001. The size of the memory bank $M$ is set to 16384 and $\tau$ is 0.07.

\begin{table}[t!]
	\caption{Performance comparison on NTU 60 under semi-supervised evaluation protocol.} \label{tab:ntu_semi}
		\centering
		\begin{tabular}{l|cccc}
			\toprule
			\multirow{4}{*}{Method} & \multicolumn{4}{c}{{NTU 60}}\\
			\cmidrule(lr){2-5}
			& \multicolumn{2}{c}{{xview}} & \multicolumn{2}{c}{{xsub}} \\
			\cmidrule(lr){2-3} \cmidrule(lr){4-5}
			&   1\% data & 10\% data  & 1\% data & 10\% data\\
			\midrule
			LongT GAN \cite{zheng2018unsupervised}   &  - & - & 35.2 & 62.0 \\
			MS$^2$L \cite{lin2020ms2l}  &  - & - & 33.1 & 65.1   \\
			HiCLR \cite{zhang2022hiclr} &50.9 &79.6 & 51.1 & 74.6 \\
      ISC \cite{thoker2021skeleton} &  38.1 &72.5 &35.7 &65.9  \\
			CMD~\cite{mao2022cmd} &  53.0 &80.2 &50.6 &75.4\\
			\textbf{PCM$^{\rm 3}$} &\textbf{53.1} &\textbf{82.8}  &\textbf{53.8}  &\textbf{77.1}  \\
			\bottomrule
		\end{tabular}
\end{table}

\subsection{Comparison with State-of-the-Art Methods}
To give a comprehensive evaluation of the generalization capacity of the proposed method, PCM$^{\rm 3}$, we conduct experiments on the following \textit{five} downstream tasks under three widely used datasets.

\noindent \textbf{1) Skeleton-based Action Recognition.} After pretraining the encoder $f(\cdot)$ on the self-supervised tasks, we utilize the learned representations to solve the skeleton-based action recognition. Specifically, two evaluation approaches are adopted, \ie,~unsupervised learning approach and semi-supervised learning approach.

\noindent $\bullet$ \textbf{Unsupervised Learning Approach} applies a fully-connected layer after the encoder, which is fixed during training. We report the top-1 accuracy results in Table~\ref{tab:linear_ntu}. On NTU datasets, PCM$^{\rm 3}$ can surpass other state-of-the-art methods notably on all protocols. Especially on NTU 120 dataset, our method shows 5+\% improvements compared to the latest methods. Remarkably, the single stream of our method can perform on par with the three streams of SOTA methods. Meanwhile, we give the results on PKUMMD II, which is a relatively small dataset but contains more noisy data in real life. PCM$^{\rm 3}$ can achieve the best results on all benchmarks, indicating the strong generalization capacity and robustness across datasets. 

\noindent $\bullet$ \textbf{Semi-supervised Learning Approach} jointly trains the encoder $f(\cdot)$ and a fully-connected layer for the recognition task. But only a portion of the labeled training data is available, \ie,~1\% and 10\%. This reflects the representation quality because a good representation can effectively avoid the over-fitting problem in training. The results are shown in Table~\ref{tab:ntu_semi}. As we can see, PCM$^{\rm 3}$ renews the state-of-the-art scores with varying proportions of available training data, indicating the strong generalization capacity.

\noindent \textbf{2) Skeleton-based Action Retrieval.} We follow the settings introduced by previous work~\cite{su2020predict}. Specifically, the K-nearest neighbors (KNN) classifier ($k$=1) is employed to the learned representations to assign action labels for the training set. 
The results on NTU 60 and NTU 120 datasets are shown in Table~\ref{tab:knn_ntu}. The proposed method achieves the best results, surpassing other state-of-the-art methods by a large margin. This indicates a highly distinguishable representation space is obtained through our method. 

	\begin{table}[t]
              \vspace{0pt}
			\caption{Action retrieval results with joint stream.}
			\centering
  \begin{tabular}{l|cc|cc}
   \toprule
    \multirow{2}{*}{Method}&\multicolumn{2}{c|}{NTU 60}&\multicolumn{2}{c}{NTU 120}\\
    & xsub&xview &xsub& xset\\
    \midrule
    LongT GAN~\cite{zheng2018unsupervised} & 39.1 &48.1 &31.5 &35.5\\  
    ISC~\cite{thoker2021skeleton}&62.5	&82.6 	&50.6 &	52.3\\
    CRRL~\cite{wang2022contrast} &60.7	&75.2	&-	&-\\
    CMD~\cite{mao2022cmd} & 70.6	&85.4  	&58.3	&60.9\\
    HaLP~\cite{shah2023halp} &65.8 &83.6 &55.8 &59.0\\
    \textbf{PCM$^{\rm 3}$} &\textbf{73.7}&\textbf{88.8} &\textbf{63.1} &\textbf{66.8} \\
    \bottomrule
\end{tabular}
  \label{tab:knn_ntu}
  \end{table}	

\begin{table}[t]
              \vspace{0pt}
		\caption{Action recognition with occlusion results. $\Delta_{\downarrow}$ represents the average performance reduction compared to that without occlusion.}
			\centering
  \begin{tabular}{l|ccc|ccc}
   \toprule
   \multirow{3.5}{*}{Method} & \multicolumn{6}{c}{{Occluded NTU 60}}\\
    \cmidrule(lr){2-7}
    &\multicolumn{3}{c|}{Spatial Occ.(\%)}&\multicolumn{3}{c}{Temporal Occ. (\%)}\\
    & xsub&xview &$\Delta_{\downarrow}$ &xsub& xview&$\Delta_{\downarrow}$\\
    \midrule
    MoCo-GRU~\cite{he2020momentum} &64.8  &72.6 &12.4&68.8  &74.8 &9.3\\
    ISC~\cite{thoker2021skeleton}&62.8	&70.6 &14.1 &68.9&76.8&7.9 \\
    CRRL~\cite{wang2022contrast} &56.8 & 61.4 &11.6  &61.0 &66.2 &7.1\\
    AimCLR~\cite{guo2022aimclr} &54.9&58.5  & 20.3&54.1&58.6 & 20.7\\
    CMD~\cite{mao2022cmd} &67.1 &72.7	&13.3  	&72.7 &79.5	&7.1\\
    \textbf{PCM$^{\rm 3}$} &\textbf{80.8}&\textbf{87.0} &\textbf{3.3} &\textbf{77.6} &\textbf{86.1} &\textbf{5.4} \\
    \bottomrule
\end{tabular}
  \label{tab:occlusion}
	\end{table}

\noindent \textbf{3) Action Recognition with Occlusion.} Occlusions are universal disruptions that constantly occurred in the real world, which can seriously affect the performance of action recognition. We transfer the learned representations from clean dataset to the action recognition task with body occlusion. A linear evaluation protocol is adopted. Following the work~\cite{song2019richly}, we use a synthetic dataset with both spatial and temporal occlusion. For spatial occlusion, we randomly masked the body parts, \eg~\textit{trunk} and \textit{right-hand}. For temporal occlusion, we randomly mask a block of frames to zeros. All masks are generated randomly with the masking ratio sampled from [0.3, 0.7]. 

The experiments are conducted under the same settings across compared methods, and the results are shown in Table~\ref{tab:occlusion}. Due to the proposed topology-based masked contrastive learning pretraining, our method can capture the discriminative structures in the distorted data and well handle the spatial occlusion. Meanwhile, this ability also extends well to temporal occlusion. As we can see, our approach showed significant improvements in both occlusion scenarios, as well as minimal performance degradation compared to that under clean data.

 \begin{table}[t]
  \centering
  \caption{Action detection results on PKUMMD Part I xsub benchmark with overlap ratio of 0.5.}
  \begin{tabular}{l|c|c}
   \toprule
    {Method}&{mAP$_{\rm a}$ (\%)}&{mAP$_{\rm v}$ (\%)}\\
    \midrule
    Randomly Initializaed &29.6 & 28.2\\
    MS$^{2}$L~\cite{lin2020ms2l}&50.9 & 49.1\\
    CRRL~\cite{wang2022contrast}&52.8 & 50.5\\
    ISC~\cite{thoker2021skeleton}&55.1 & 54.2\\
    CMD~\cite{mao2022cmd} & 59.4  &59.2\\
    \textbf{PCM$^{\rm 3}$}&\textbf{61.8} &\textbf{61.3} \\
    \bottomrule
\end{tabular}
  \label{tab:detection}
\end{table}

\noindent \textbf{4) Skeleton-based Action Detection.} Following the settings in~\cite{liu2020pku,chen2022hierarchically}, we evaluate the detection performance under the PKUMMD I dataset, to demonstrate effectiveness for the short-term frame-level discrimination task. We attach a linear classifier (fully-connected layer) to the encoder and finetune the whole model to predict the label of each frame. The encoder is pretrained on NTU 60 xsub dataset and then transfers to the PKUMMD Part I xsub dataset. We adopt the mean average precision of different actions (mAP${\rm_a}$) and different videos (mAP${\rm_v}$) with the overlapping ratio of 0.5 as the evaluation metrics. The results are shown in Table~\ref{tab:detection}. First, we can see the existing contrastive learning methods can largely boost the detection performance compared with the randomly initialized model. It indicates the learned sequence-level representations can benefit the frame-level task to some extent. Besides, our method further improves the state-of-the-art scores owing to the synergetic modeling of the joint-level and sequence-level features.

\begin{table}[t]
  \centering
  \caption{The results on motion prediction task.}
  \begin{tabular}{c|c|c|c|c|c}
   \toprule
    \multirow{2}{*}{Method}&\multirow{2}{*}{Random}&{CRRL}&{ISC}&{CMD}&\multirow{2}{*}{PCM$\rm ^{3}$}\\
    & &\cite{wang2022contrast} & \cite{thoker2021skeleton} & \cite{mao2022cmd} & \\
    \midrule
    MPJPE (mm) &108.5 &104.9 &144.2 &145.0 &\textbf{101.7}\\
    \bottomrule
\end{tabular}
  \label{tab:motion_prediction}
\end{table}

\begin{table*}[t]
  \centering
  \caption{Ablation study on the synergy of the contrastive learning and masked skeleton prediction on different downstream tasks. \textit{Multi-Task} stands for simply combining the two tasks in a multi-tasking manner.}
  \begin{tabular}{l|c|c|c|c|c}
   \toprule
    \multirow{2}{*}{Method}&{Recognition}&{Retrieval}&{Recognition \textit{w} Occlusion}&{Detection}&{Motion Prediction}\\
    & Accuracy (\%)&Accuracy (\%)&Accuracy (\%) & mAP$\rm _a$ (\%)& MPJPE \\
    \midrule
    Masked Prediction &14.7 &64.4 &11.7 &47.0 &102.8  \\
    Contrastive Learning & 87.3 &84.5  &76.6 &58.4 &147.3 \\
    Multi-Task & 87.5 &85.1 &77.4 &59.7 &103.9 \\
    Ours & \textbf{90.4} &\textbf{88.8} &\textbf{87.0} & \textbf{61.8} &\textbf{101.7} \\
    \bottomrule
\end{tabular}
  \label{tab:single_multi}
\end{table*}

\begin{table}[t]
  \centering
  \caption{Ablation study on the different positive samples and the distillation loss (without any prompt applied).}
  \begin{tabular}{cc|cc|c|c}
   \toprule
   \
   $s_{intra}$ &$s_{inter}$  &$s_{mask}$  &$s_{predict}$ & $\mathcal{L}_{KL}$ &Acc. (\%) \\
    \midrule
    \checkmark & & & & & 84.7 \\
    
    \checkmark &\checkmark & & & & 87.5\\
    \midrule
    \checkmark &\checkmark & & &\checkmark &88.2 \\
    \checkmark &\checkmark & &\checkmark &\checkmark & 89.7 \\
    \checkmark & &\checkmark &  &\checkmark & 89.3 \\
    \checkmark &\checkmark & \checkmark&\checkmark &\checkmark  &\textbf{90.0} 
    \\
    \bottomrule
\end{tabular}
  \label{tab:abl_main_design}
\end{table}

\begin{table}[t]
			\caption{Ablation study on the different types of prompts.}
			\centering
			\setlength{\tabcolsep}{6pt}
			  \begin{tabular}{cc|cc}
   \toprule
    Domain-& Task-&Recognition (\%) &Detection (\%)\\
    \midrule
    & & 90.0 & 60.8\\
    \checkmark &  &90.3 & 61.0\\
    \checkmark & \checkmark &\textbf{90.4} &\textbf{61.3} \\
    \bottomrule
\end{tabular}
  \label{tab:diff_prompt}
	\end{table}	
 
\noindent \textbf{5) Motion Prediction.}
Following the previous work~\cite{chen2022hierarchically}, we give the results of the motion prediction task, which is a dense prediction downstream task. We use the decoder in~\cite{martinez2017human} after $f(\cdot)$ and follow the short-term motion prediction protocol~\cite{martinez2017human}. As shown in Table~\ref{tab:motion_prediction}, our method achieves the best results in terms of the MPJPE metric. Although previous contrastive learning-based methods show good performance in a high-level recognition task, most of them show adverse effects on motion prediction compared with the randomly initialized method. It is because they only focus on the high-level information and ignore the joint-level feature modeling. In contrast, our method extracts the joint-level and sequence-level semantic features and effectively boosts the dense-prediction downstream task performance.

\subsection{Ablation Study}
In this part, we give a more detailed analysis of the proposed method. The results are reported on the action recognition task under linear evaluation, using NTU 60 dataset xview protocol by default. 

\begin{table}[t]
  \centering
  \caption{Ablation study on the masking strategy.}
  \begin{tabular}{l|c|c}
   \toprule
    {Mask Ratio}&{{Topology-based}}&{{Random}}\\
    \midrule
    20\% & 89.8 & 89.0\\
    40\% & 90.2 & 89.5\\
    60\% & \textbf{90.4} &\textbf{89.7} \\
    80\% & 90.1 &{89.5}\\
    \bottomrule
\end{tabular}
  \label{abl:mask}
\end{table}

\noindent \textbf{1) Effect of masking strategy.}
We utilize the \textit{topology-based} masking strategy in our framework to construct more challenging corrupted data views.
In Table~\ref{abl:mask} we report the linear evaluation results for action recognition task because it is widely used for evaluation of other representation learning methods. Compared with \textit{Random Mask} strategy, 
the model shows better results, indicating a more distinguishable feature space.
Meanwhile, more realistic occluded data are simulated by this masking strategy, which are often continuous in spatial-temporal dimension. We set mask ratio to 0.6.

\noindent \textbf{2) Analysis of the synergetic design between the two tasks.}
We first analyze the performance under different downstream tasks when adopting single paradigm, \ie,~masked prediction and contrastive learning, or combinations of them, in Table~\ref{tab:single_multi}. Only employing masked prediction task cannot generate highly distinguishable feature space, resulting in poor performance in linear recognition task. Meanwhile, the contrastive learning mainly modeling high-level features and the learned prior representations are not beneficial for motion prediction. Therefore, to combine the merits of both paradigms, we can directly perform the multi-task learning. Owing to the well-designed contrastive learning and masked prediction pipelines, naive multi-task method can achieve a decent performance. However, it ignores the connection between the two tasks and only shows 
mediocre improvements on the recognition task. In contrast, our design utilizes the synergy between the two tasks, and further improve the performance and generalization capacity, proving the effectiveness of the proposed method.

Next, we elaborate on the separate effect of our synergetic designs. As shown in Table~\ref{tab:abl_main_design}, the proposed inter- and intra- skeleton transformations can well boost the contrastive learning performance. Meanwhile, the masked views and predicted views in the masked prediction training as the positive pairs yield 1.1\% and 1.5\% improvement, respectively. The relational distillation loss is found effective and further improve representation quality. 

Finally, we show the effect of propagating the gradients of contrastive branch to masked prediction here, \textit{80.5} with gradient vs. \textit{80.0} without gradient, that is stopping and separating the gradients of two branches. The gradient information from contrastive branch can serve as high-level semantic guidance, which boosts the masked prediction training and yields more informative positive samples. 

\noindent \textbf{3) Effect of the dual-prompted pretraining strategy.} We show the effect of different types of prompts in Table~\ref{tab:diff_prompt}.
As we can see, the proposed domain-specific and task-specific prompts are effective for the representation learning, which can learn the domain- and task-specific knowledge and serve as explicit guidance in-training. This assists the model to discriminate the domain and task identity, reducing the interference and ambiguity when the model learns from multi-view data and multiple tasks. When the two types of prompt are utilized, the model achieves the best results.

\section{Conclusion}
We propose a novel framework called prompted contrast with masked motion modeling, PCM$^{\rm 3}$, which can effectively learn meaningful representations by exploring the mutual collaboration between contrastive learning and masked prediction tasks. Specifically, the novel views in masked prediction training are utilized as the positive samples for contrastive learning. Meanwhile, contrastive learning provides semantic guidance for masked prediction in turn by propagating the gradients to the prediction decoder. Furthermore, we introduce dual-prompt multi-task pretraining strategy to provide explicit guidance. Extensive experiments are conducted, demonstrating the superior performance and promising generalization capacity of our method.

\bibliographystyle{ACM-Reference-Format}
\balance
\bibliography{sample-base}

\end{document}